\documentclass[10pt,twocolumn,letterpaper]{article}
\usepackage{cvpr}
\usepackage{times}
\usepackage{epsfig}
\usepackage{graphicx}
\usepackage{amsmath}
\usepackage{amssymb}
\usepackage{enumitem}
\usepackage{subcaption}
\usepackage[export]{adjustbox}
\usepackage{pifont}
\usepackage{lipsum}
\usepackage[square,sort,comma,numbers]{natbib}
\UseRawInputEncoding
\usepackage{tabularx} 

\usepackage[pagebackref=true,breaklinks=true,letterpaper=true,colorlinks,bookmarks=false]{hyperref}

\cvprfinalcopy 

\def\cvprPaperID{****} 

\usepackage[table]{xcolor}
\definecolor{green}{RGB}{14,0,18}

\usepackage{mathbbol}
\usepackage{amssymb}
\usepackage{pbox}
\usepackage{multirow}
\begin{document}

\title{Object Detection in Aerial Images: What Improves the Accuracy?}
\author{Hashmat Shadab Malik\\
{\small hashmat.malik@mbzuai.ac.ae}
\and
Ikboljon Sobirov\\
{\small ikboljon.sobirov@mbzuai.ac.ae}
\and
Abdelrahman Mohamed\\
{\small abdelrahman.mohamed@mbzuai.ac.ae}
}
\maketitle

\begin{abstract}
Object detection is a challenging and popular computer vision problem. The problem is even more challenging in aerial images due to significant variation in scale and viewpoint in a diverse set of object categories. Recently, deep learning-based object detection approaches have been actively explored for the problem of object detection in aerial images. In this work, we investigate the impact of Faster R-CNN for aerial object detection and explore numerous strategies to improve its performance for aerial images. We conduct extensive experiments on the challenging iSAID dataset. The resulting adapted Faster R-CNN obtains a significant mAP gain of 4.96\% over its vanilla baseline counterpart on the iSAID validation set, demonstrating the impact of different strategies investigated in this work. 
\end{abstract}

\section{Introduction}
Object detection (OD), one of the computer vision tasks, poses its own challenges on top of mere identification or localization tasks \cite{survey}. It is a task of classification and localization of objects in an image. It has gained sufficient fame to be a major field of research in computer vision on account of its efficacy and wide use in real life applications. To make the task even more challenging, OD in aerial images is emerging as a new task in which minute objects are generally of interest of detection. 

Traditionally, machine learning was the weak solution to OD tasks. An ensemble of hand-crafted feature extractors, such as histogram of gradients \cite{hog}, were commonly in use. Recent strides in deep learning (DL) have proved its viability in OD tasks, yielding promising and applicable results. Not only did their advantage of automating the feature extraction process make the network construction effortless, they also produced a substantial boost in results. Therefore, the primary approach to tackling the OD tasks at present is DL-driven models.

In this work, we apply one of the most popular architectures for OD, Faster R-CNN \cite{Faster_RCNN}, and make numerous modifications on top to reach higher results. Our contributions are as follows:
\begin{itemize}
    \itemsep-0.5em 
    \item Examining a new backbone to the Faster R-CNN network
    \item Exploring several data augmentations that contribute to an increase in results
    \item Proposing a deeper region proposal network that integrates a spatial and channel squeeze and excitation block
    \item Investigating soft non-max suppression technique in the Faster R-CNN network 
\end{itemize}

\begin{figure}
    \centering
    \includegraphics[width=0.45\textwidth]{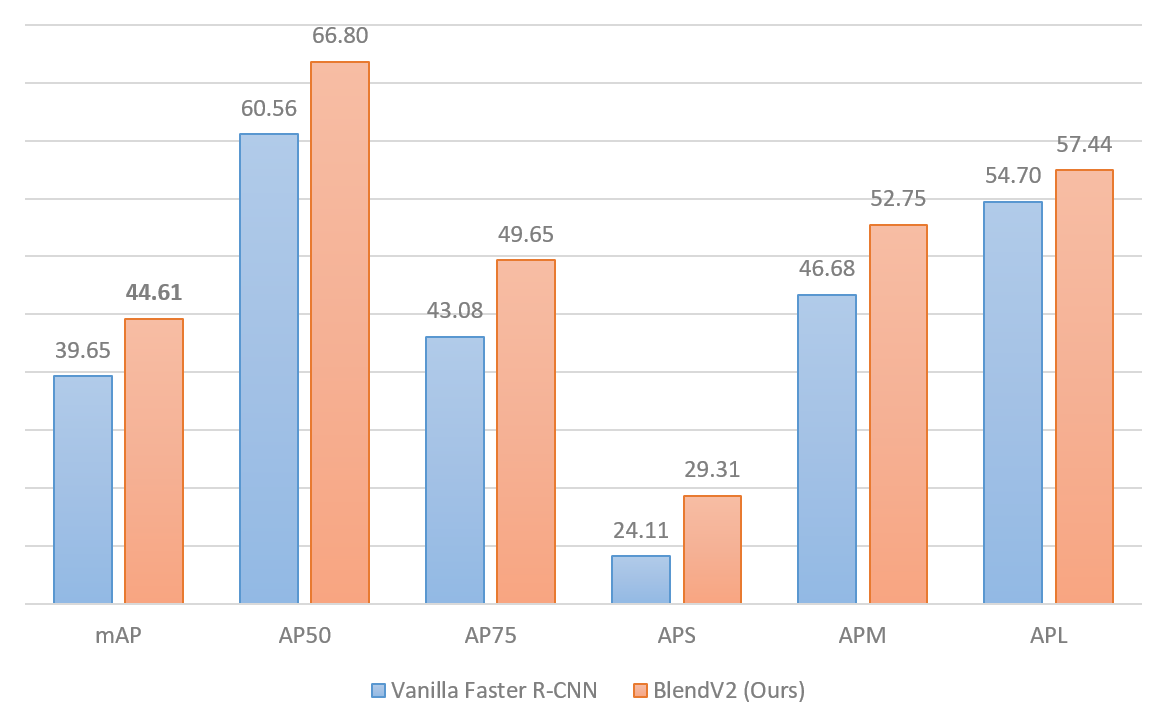}

   \caption{The figure shows the results of the improvements introduced on top of the vanilla Faster R-CNN.}
    \label{intro_results}
\end{figure}

\section{Related Work}
Object detection task of computer vision on its own is challenging, and therefore, is attractive to many researchers in the field. It has been extensively studied with the use of natural images with promising models and results. Aerial images is newly emerging as an area of interest. It poses a few more issues on top of traditional OD, such as small scales and size irregularities in images. This section first highlights the common architectures used for OD, and second, reviews recent papers that tackled the OD task with natural as well as aerial images.

\subsection{Architecture Review}
Starting in early years of OD, traditional machine learning approaches were implemented to extract features; advances in deep learning has overthrown the traditional methods, offering multiple advantages, including higher accuracy and precision. At an eagle view on current approaches, two categories can be extracted: one-stage OD and two-stage OD. Both categories are heavily in use today, with one-stage being much faster in inference time and two-stage yielding higher results. Of the many, YOLO family \cite{yolo}\cite{yolo2}\cite{yolo3}\cite{yolo4}\cite{yolo5} is one of the most go-to methods in object detection for their continuous improvements in each newer version, both in time and accuracy. R-CNN family \cite{rcnn}\cite{fast_rcnn}\cite{Faster_RCNN} comes toe-to-toe with the YOLO counterpart, with its higher results that is superior to YOLO. To be more specific, Faster R-CNN \cite{Faster_RCNN} is the last version of R-CNN. It is a single and unified network for object detection, and one of the most popular two stage detectors. It builds upon its predecessor Fast R-CNN \cite{fast_rcnn} by introducing two new concepts: anchor boxes and region proposal network (RPN).  


\begin{figure}
    \centering
    \includegraphics[width=0.22\textwidth]{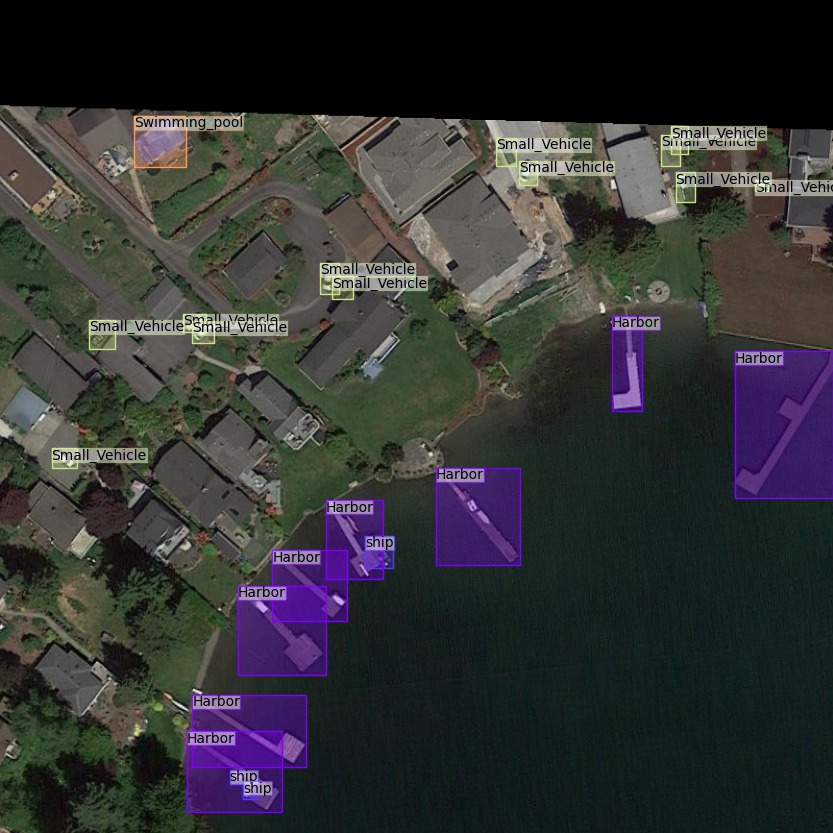}
    \includegraphics[width=0.22\textwidth]{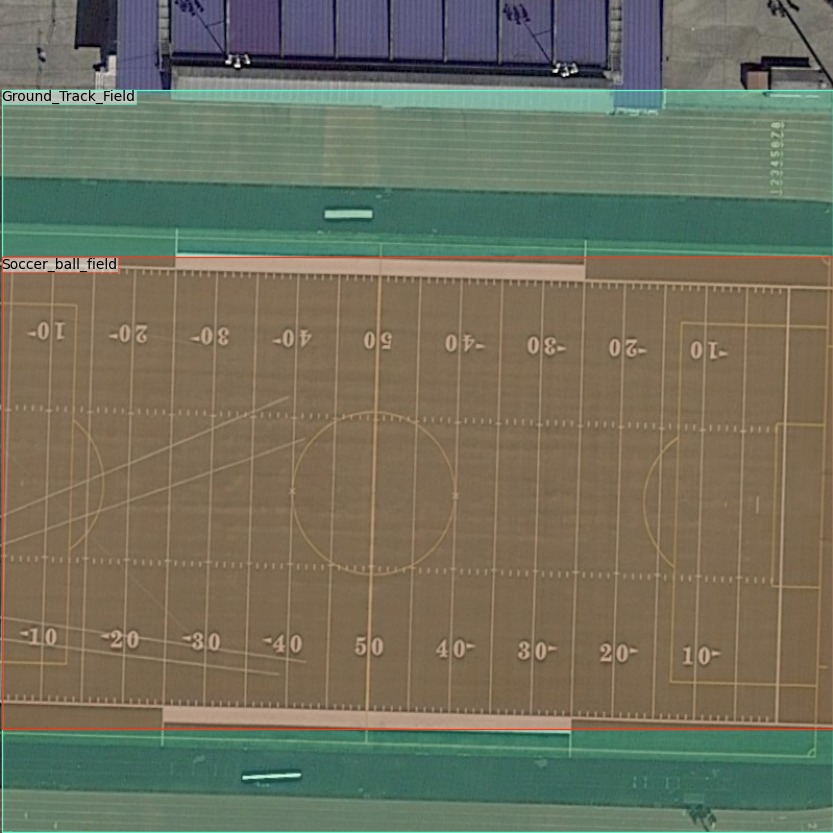}
    \includegraphics[width=0.22\textwidth]{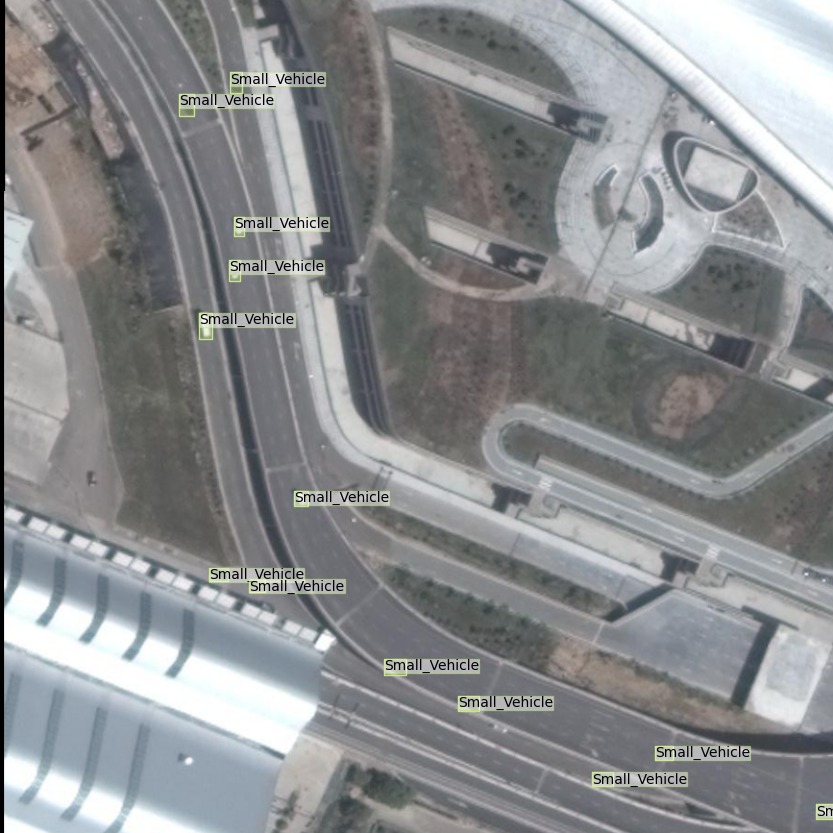}
    \includegraphics[width=0.22\textwidth]{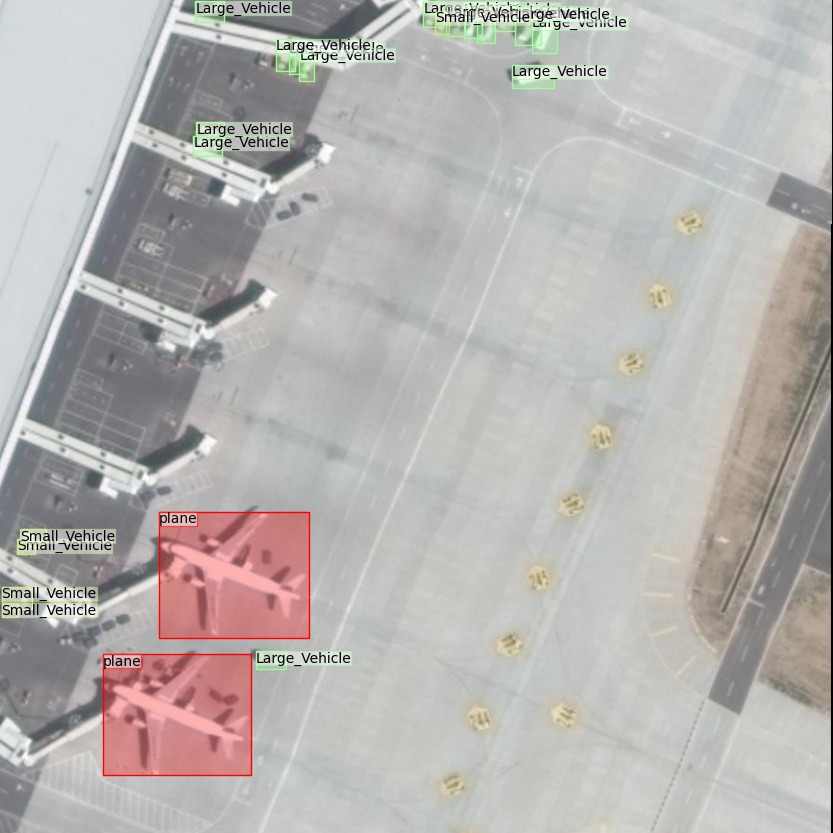}

    \caption{The figure shows a few samples from the iSAID dataset. Note that some images contain only one category, while others contain several within the same image.}
    \label{samples}
\end{figure}

\subsection{Recent Approaches}
Building on top of the common architectures that yield promising results, multiple papers tried to make improvements in accuracy and/or time. In \cite{wei}, the authors improve upon the vanilla SSD \cite{ssd} to include multi-scale context information to reach higher accuracy. They use dilated and deconvolution layers in the context layers, and claim that they are the two variants of CSSD - a shorthand for context-aware single-shot multibox object detector. Further improvements on SSD were introduced in \cite{fahadsug}\cite{fahadsug2}. D2Det method presented in \cite{d2det} is a two stage approach for object detection and instance segmentation in general. They validate their approach on MS COCO \cite{mscoco}, UAVDT \cite{uavb} and iSAID \cite{isaid} for object detection and instance segmentation, outperforming other state-of-the-art methods at the time.

Several approaches have been proposed for object detection with aerial images. Igor et al. \cite{igor} present a simple CNN-based architecture to automate image classification and OD in aerial images. Sommer et al. \cite{sommer} implement Fast R-CNN and Faster R-CNN to detect vehicles in aerial images, and validate their results on two publicly available datasets. In \cite{fan}, ClusDet algorithm is proposed to detect objects in the same domain. To address the issues of small objects in aerial images and the non-uniformity in data, they incorporate a cluster proposal network to extract cluster regions, a scale estimation network and a separate detection network. To make the task even more challenging, \cite{dardet} tackle the rotated OD (i.e. the bounding boxes are not axes aligned). They propose a dense anchor-free rotated object detector to do this task, with five parameters of prediction.




\section{Datasets}
\begin{figure}
    \centering
    \includegraphics[width=0.45\textwidth]{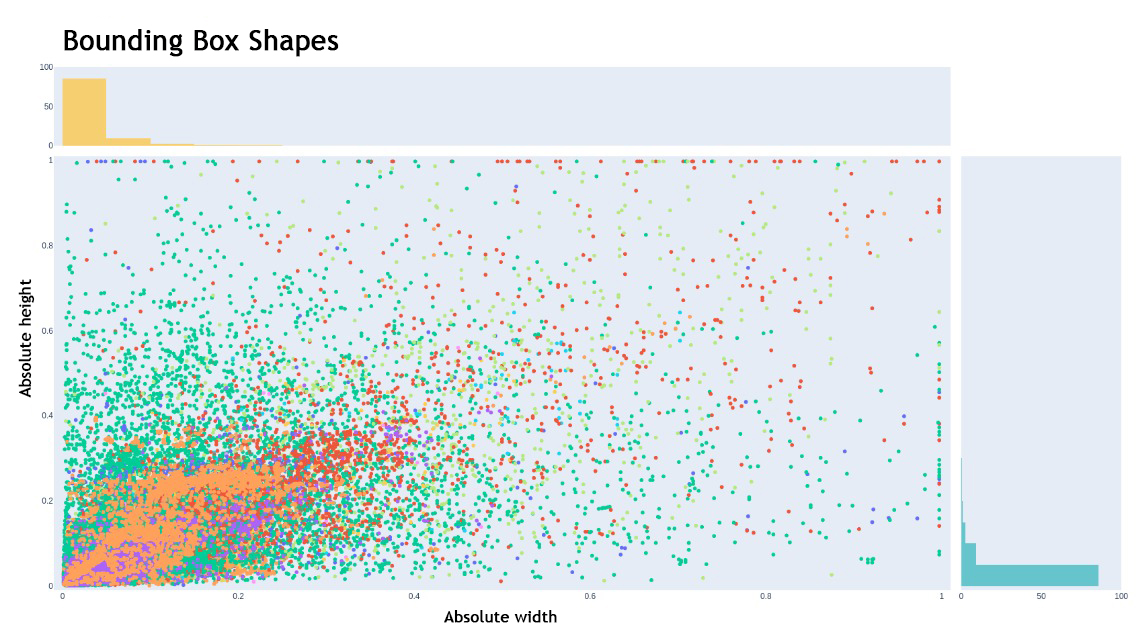}
    \caption{The figure shows height-to-weight correlation of bounding boxes of objects in the dataset. Each color represents a different category. Side bars are the proportion of how many instances of bounding boxes lie within that size range.}
    \label{data_analysis}
\end{figure}

Natural scene datasets as ImageNet \cite{imagenet}, Pascal VOC \cite{pascalvoc}, CityScapes \cite{city} and MSCOCO \cite{mscoco} are the largest scale datasets available for use in deep learning (DL). However, their use is limited to upward orientation of the images, meaning that they are generally captured from a side view. When the viewpoint changes, for example, to a top view, the generalizability of the DL methods starts to deteriorate. In aerial images, the challenges to traditional imaging accumulate with a few more such as irregularities in shapes and orientations, object instances in higher densities, or large aspect and scale variations \cite{isaid}. There are tenth of aerial, satellite and Google Earth datasets slowly emerging since 2008 up to now, and is still under much attention as a research area. Although the prevalence of such datasets is high, the total number of categories in each is usually limited to only a few. The datasets with the most number of categories and instances of these categories and the most popular ones are DIOR \cite{dior}, DOTA \cite{dota}, xVIEW \cite{xview} and iSAID \cite{isaid}, and iSAID dateset is used for the current work.

\begin{figure} 
    \centering
    \includegraphics[width=0.4\textwidth,height=0.35\textheight,keepaspectratio]{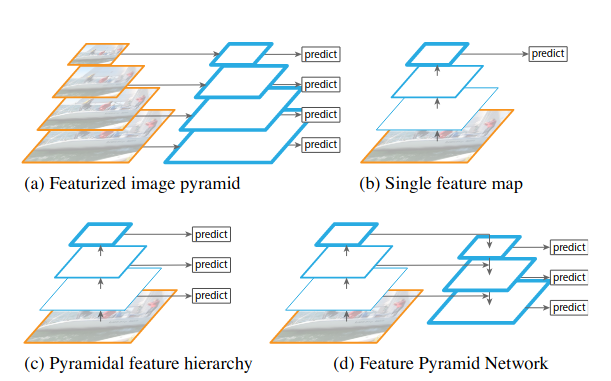}
    \caption{The figure shows different methods (a, c, d) to extract multi-scale information using different scales of feature maps, while also showcasing the simple single feature map (b) used in vanilla Faster R-CNN (Adapted from \cite{fpn_paper}).}
    \label{Pyramid_networks}
\end{figure}

\begin{figure}
    \centering
    \includegraphics[width=0.4\textwidth,height=0.35\textheight,keepaspectratio]{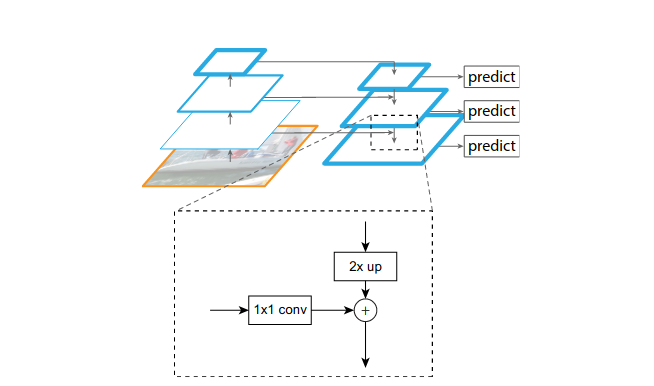}
    \caption{The figure shows the addition block showcasing how multi-scale feature maps are added in the FPN (Adapted from \cite{fpn_paper}).}
    \label{Addition_block}
\end{figure}

Unlike other counterparts, iSAID dataset is the first benchmark in instance segmentation task of computer vision using aerial images. 2806 high-resolution images that includes 655451 object instances for 15 categories are provided in the dataset. The categories are ship, storage tank, baseball diamond, tennis court, basketball court, ground track field, bridge, large vehicle, small vehicle, helicopter, swimming pool, roundabout, soccer ball field, plane and harbour. iSAID dataset images are of large resolutions and the DL models cannot handle such large quality image, thus, they are preprocessed. Specifically, patch sizes of \texttt{800x800} are extracted from the original images, now equalling 28029 and 9512 images for training and validation respectively. Figure \ref{samples} depicts some samples from the dataset with the bounding boxes drawn on the corresponding category images. As can be seen, the dataset contains images with only a single category, as in Figure \ref{samples} lower left (only small vehicles), and several categories within the same image, as in upper left (harbour, small vehicle, swimming pool and ship) or the other two on the right. It is also noteworthy that there are multiple instances of the same category within the same image, which is common in the dataset.

The authors list down a few differentiating characteristics of the dataset to contrast it to other ones, claiming that the dataset has:

\begin{figure}
    \centering
    \includegraphics[width=0.4\textwidth]{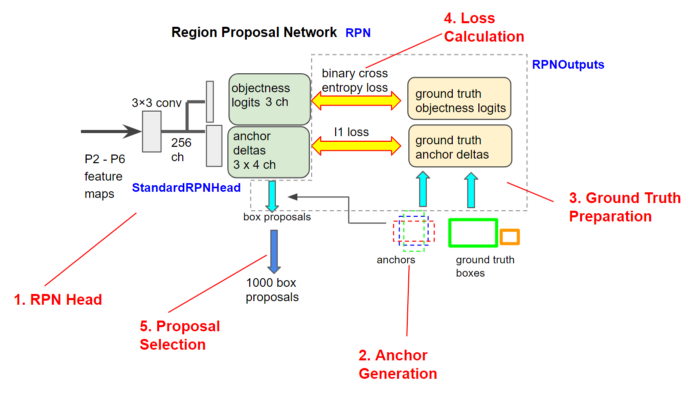}
    \caption{The figure shows the design of the RPN block in the baseline (Adapted from \cite{RPN_figure}). At each position of the feature map, objectness score of all the anchors (3 by default) at that position as well as there predicted shift is predicted.}
    \label{RPN_Block}
\end{figure}

\begin{figure}
    \centering
    \includegraphics[width=0.4\textwidth,height=0.35\textheight,keepaspectratio]{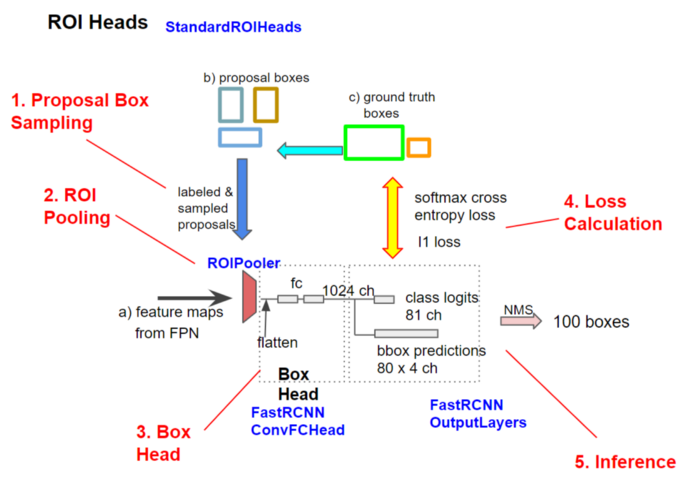}
    \caption{The figure shows the design of the ROI block in the baseline (Adapted from \cite{RPN_figure}). Each proposed region from the RPN is cropped from the feature map, then an ROIPooler, in our case ROIAlignV2, is used to make all cropped regions have the same spatial size. Then each cropped region is passed through a fully connected layer into two branches, one to output the classes prediction and the other to predict the bounding box shift.}
    \label{ROI_Block}
\end{figure}

\begin{figure*}
    \centering
    \includegraphics[width=1.0\textwidth,height=0.3\textheight,keepaspectratio]{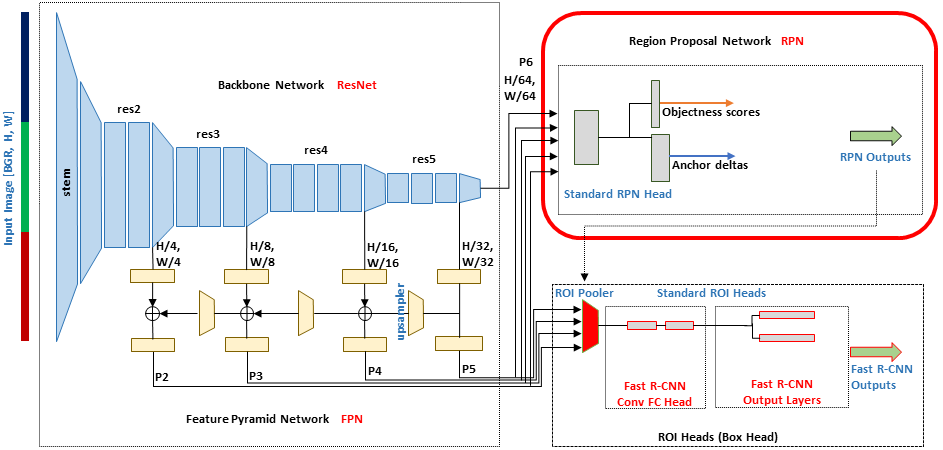}
    \caption{The figure shows the baseline architecture of Faster R-CNN with FPN-based ResNet backbone. Features extracted at different layers of the backbone are passed to the RPN module as well as ROI head module. RPN produces proposals used in the RIO head module to yield final results of the object detection task.}
    \label{Faster_RCNN}
\end{figure*}

\begin{itemize}
    \itemsep-0.5em 
    \item a significant quantity of high spatial resolution images
    \item fifteen categories that are common and important in nature 
    \item a significantly increased number of object instances in each image
    \item a considerable number of labeled instances per image
    \item large scale variations of objects (small, medium and large), existing in the same picture in many cases
    \item data imbalance with respect to objects in images reflecting real-life aerial imaging
    \item small sized objects with appearances in ambiguity that requires context to get resolved
    \item high precision in instance level annotations by experts that are cross-validated by other professionals who followed guidelines for annotations. 
\end{itemize}

To understand the dataset in further details, bounding box information was extracted. The object dimensions are first normalized for this analysis. We plot in Figure \ref{data_analysis} height and width of each object bounding box in a correlated fashion. Each color represents a different category. As can be seen in the side bars, most heights and widths (about 80 percent) lie within the range of 0.2 and 0.2 respectively. It indicates that most objects in the images are relatively small, and there are a few that occurs at the highest dimensions.

\section{Methods}
\subsection{Baseline}
Our baseline for this project is a variation of two-stage object detection model - Faster R-CNN. In the provided dataset, we have objects which vary significantly in scale. To accurately capture all the objects using vanilla Faster R-CNN, we need to have a very high variation in the anchor scales chosen to generate region proposals on the single feature map scale. This task of identifying objects at different scales, and specifically small objects, is very challenging. Furthermore, using images at different scales to overcome this issue is constrained by the memory as well as time consumption while training.

Taking the above statements into consideration, we chose Feature Pyramid Network (FPN) based Faster R-CNN \cite{fpn_paper} as our baseline model. This network (as shown in Figure \ref{Faster_RCNN}) exploits the hierarchical property of convolution neural networks (in our case, a ResNet backbone \cite{Resnet}) to extract features at multiple levels rather than taking images at multiple scales. Multi-scale feature maps generated in this manner provides the model with better quality information than the single-level feature map used in the vanilla Faster R-CNN. Figure \ref{Pyramid_networks} (d) shows the basic design of the FPN backbone used in our baseline. It can be seen in Figure \ref{Pyramid_networks} (d) that predictions are made independently on each level, while also reusing deep features learned by the subsequent layers. The smaller/deeper resolution feature maps are up-sampled (via nearest neighbors) before adding them (element-wise) to larger resolution feature maps as shown in Figure \ref{Addition_block}.

After getting features at different levels, similar to vanilla Faster R-CNN, Region Proposal Network (RPN) \cite{Faster_RCNN} is used as a class-agnostic object detector. The RPN network is originally used on top of a single-scale feature map, but in our case, it will be used on all the multi-scale feature maps generated by our FPN backbone ($P2-P6$). For each feature map, a $3\times3$ convolution layer, followed by a separate $1\times1$ convolution for the objectness score and bounding box regression is applied. As shown in Figure \ref{RPN_Block}, all the feature maps are fed into the RPN block sequentially; the higher resolution feature maps are utilised to generate proposals for small objects while as lower resolution feature maps capture proposals for larger objects. Each scale is used to predict objectness score and anchor deltas for different anchor sizes and at different strides laid on the original image; e.g. [32, 64, 128, 256, 512] anchor sizes with strides [4, 8, 16, 32, 64] corresponding to the features maps $P2-P6$. At each scale, three different aspect ratios of the anchors are used to get good proposals for objects of different shapes. The size of the anchors used in the model depends on the size of the objects in the dataset, and thus needs to be tuned to get the best proposals. 
\\
\indent After this stage, top-\textit{k} proposal are selected based on the proposals with the highest objectness score at each scale and non-maximum suppression to remove overlapping proposals. These proposals are then utilised by the ROI block to crop the corresponding regions of interests (ROIs) from the feature maps $P2-P6$. Different methods such as ROI Pooling, ROIAlign and ROIAlignV2 were used to crop the feature maps based on the generated proposals and then resize them to the same size. The ROIs generated in this fashion are then passed to a sequence of fully connected layers as shown in Figure \ref{ROI_Block} to classify the object within the ROIs and fine-tune the position of the bounding box.

\begin{figure}
    \centering
    \includegraphics[width=0.35\textwidth,height=0.2\textheight,keepaspectratio]{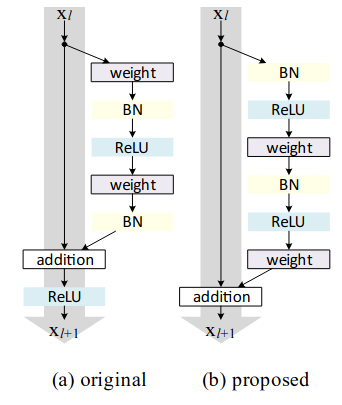}
    \caption{The figure shows the residual flow of input (and the gradient back) in the proposed residual block (b). Note that the straight line is the skip connection here. Adapted from \cite{Resnetv2}.}
    \label{Residual Unit}
\end{figure}

\subsection{Proposed Modifications on the baseline}
\subsubsection{Changing Backbone}

In the baseline model, ResNet-101 \cite{Resnet} was used as the backbone model from which we get the multi-scale features used in the subsequent modules of the network. This backbone is comprised of residual blocks which help much better flow of gradients in the backward pass, leading to better convergence for deep neural networks. In \cite{Resnetv2}, a new variant of residual unit was proposed that achieves faster error reduction and a lower training loss. Figure \ref{Residual Unit} shows the changes done in the residual unit. The authors were able to demonstrate through ablation experiments the smooth propagation of information while training deep neural networks using the proposed residual unit. It also uses group normalization and weight standardization that is useful when the batch size is small. This backbone is pretrained on ImageNet21k. In this context, we will be use the newer version of ResNet-101, known as ResNetv2-101 and analyze the change in the evaluation metric.

\begin{figure}
    \centering
    \includegraphics[width=0.35\textwidth,height=0.2\textheight,keepaspectratio]{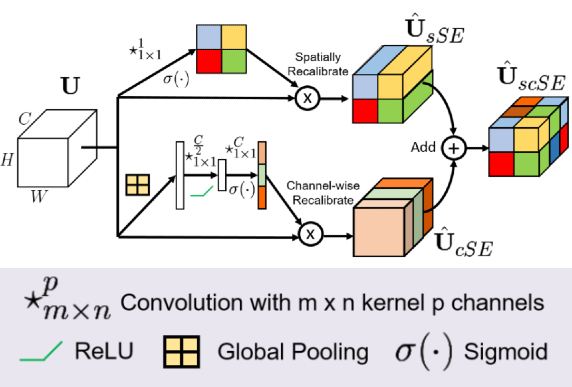}
    \caption{The figure shows the spatial and channel squeeze \& excitation block. The upper flow shows the spatial squeeze and excitation, and the lower path shows the channel squeeze and excitation, both of which are then combined together. Adapted from \cite{DBLP:journals/corr/abs-1803-02579}.}
    \label{scSE}
\end{figure}

\subsubsection{Deep-RPN block}
In order to improve the RPN block to learn more meaningful features for generating better proposals for object and background class, we introduced squeeze and excitation (SE) modules \cite{DBLP:journals/corr/abs-1803-02579} in the RPN block. We use a deeper RPN-head with feature maps recalibrated using concurrent spatial and channel squeeze excitation (scSE) modules to emphasize useful spatial regions and channel  present in the feature maps. Using spatial excitation in addition to channel excitation is more useful in our case as having pixel-wise information helps localising objects for better proposals. Figure \ref{scSE} shows the architectural design for an scSE module.


\subsubsection{Data Augmentation}
It is common that data augmentations contribute to the improvement in results, fabricating additional data for the network to learn. With that in mind, distinct sets of generic augmentations were assessed for the given task. Resizing shortest edge of objects, horizontal flip, vertical flip, rotation at 90 degrees, brightness adjustment, and saturation were all the different transforms we experimented with. To be accurate, various combinations of these augmentations were applied on the data. Supplementary Table \ref{tab:data_augs} lists all the combinations examined in this work.

\subsubsection{Soft-NMS}
Due to its suppressing criterion, non-maximum suppression (NMS) is not suitable for detecting objects that are in close proximity to each other. As it takes only the bounding box with the maximum objectness score and suppresses any other box that is above or equal a certain intersection over union (IoU) value $N_t$ with the object.

{\small
\begin{table*}\centering
  \small
\begin{minipage}{0.65\textwidth}
    \begin{tabular}{l|ccc|ccc}
      Method & AP & $AP_{50}$ & $AP_{75}$ & $AP_{S}$ & $AP_{M}$ & $AP_{L}$  \\ %
      \hline
      Baseline+NMS  &    36.04	&	56.629&	39.293&	20.88&	43.293&	48.695 \\ 
      Baseline+S-NMS  &  36.397	&	56.608&	39.986&	21.139&	\textbf{43.760}&	49.17 \\
      \hline
        Baseline+NMS+A1 & 35.788&		56.875&	38.807&	21.044&	42.645&	48.107\\ 
      Baseline+S-NMS+A1  &  36.53&		57.039&	40.11&	21.309&	43.386&	49.331 \\
      \hline
        Baseline+NMS+A2 & 35.841&		56.931&	38.888&	20.99&	42.757&	47.601\\ 
      Baseline+S-NMS+A2  &  \textbf{36.644}	&\textbf{	57.272}&	\textbf{40.238}&	\textbf{21.395}& 43.726&	\textbf{49.586} \\
      \hline
    \end{tabular}\vspace{-0.2cm}
    \caption{The table shows results of using the baseline with NMS vs Soft-NMS at various anchor sizes. Here, S-NMS stands for Soft-NMS; $A1$ and $A2$ correspond to [16, 64, 128, 256, 512] and [8, 64, 128, 256, 512] anchor sizes respectively.}\vspace{0.2cm}
  \label{tab:nms}
\end{minipage}
\end{table*}

\begin{table*}\centering
  \small
\begin{minipage}{0.65\textwidth}
    \begin{tabular}{l|ccc|ccc}
      Method & AP & $AP_{50}$ & $AP_{75}$ & $AP_{S}$ & $AP_{M}$ & $AP_{L}$  \\ %
      \hline
      Baseline-ResNet101 & 39.65	&	60.557&	43.079&	24.11&	46.679&	54.701 \\ 
      Baseline-ResNetv2-101  &  \textbf{41.977}	&\textbf{	64.562}&	\textbf{45.931}	&\textbf{27.555}&	\textbf{49.479}&	\textbf{55.971} \\
      \hline
    \end{tabular}\vspace{-0.2cm}
    \captionof{table}{The table shows the comparison of results of using different backbones. This is to show the effect of the ResNetv2 backbone stand-alone, without any further modifications applied. Note that both backbones use 101 layers in ResNet.}
  \label{tab:backbone}
\end{minipage}
\end{table*}

\begin{table*}\centering
  \small
\begin{minipage}{0.73\textwidth}
    \begin{tabular}{l|ccc|ccc}
      Method & AP & $AP_{50}$ & $AP_{75}$ & $AP_{S}$ & $AP_{M}$ & $AP_{L}$  \\ %
      \hline
      Baseline-ResNetv2 + A2 & 43.085&	66.34&	47.585&28.787&	49.962&	55.218 \\ 
      + \# of DB at 1000  &  43.715	&67.675	&48.118	&29.33	&50.572&	55.853\\
      + NMS thresh at 0.6 & 43.717	&67.432	&48.261&	29.346&	50.558&	55.657\\
      \hline
      Baseline-ResNetv2+Deep-RPN +A2& 43.234 & 66.330 & 47.486 & 28.925 &50.211 & 56.937 		 \\ 
      + \# of DB at 1000  &   43.928 &\textbf{67.766} &48.098 &29.551 & 50.942& \textbf{57.354} 		 \\
      + NMS thresh at 0.6 &\textbf{43.962}& 67.544&\textbf{48.309} &\textbf{29.584}&\textbf{50.996}&57.260			\\
      \hline

    \end{tabular}\vspace{-0.2cm}
    \captionof{table}{The table shows the results of using Deep-RPN with spatial and channel squeeze and excitation blocks. Here, $A2$ correspond to [8, 64, 128, 256, 512] anchor sizes; The second and third rows of both table sections mean that the previous model was supplemented with setting the detection boxes at 1000 and then setting the NMS threshold at 0.6.}\vspace{0.2cm}
  \label{tab:deeprpn}
\end{minipage}
\end{table*}

\begin{table*}\centering
 
\begin{minipage}{0.73\textwidth}\centering
    \begin{tabular}{l|ccc|ccc}
      Method & AP & $AP_{50}$ & $AP_{75}$ & $AP_{S}$ & $AP_{M}$ & $AP_{L}$  \\ %
      \hline
       BlendV1& 43.636 & 66.965 &48.267 &29.821 &50.957 & 55.460 				 \\ 
      + \# of DB at 1000  &   44.390 &\textbf{68.485} & 48.959 &30.497 & 51.626 &56.021 			 		 \\
      + NMS thresh at 0.6 &44.457 & 68.245& 49.182 &\textbf{30.574}&51.676 & 56.077					\\
      \hline
      BlendV2&  44.135 & 66.038 & 49.110 & 29.010& 52.102 &56.983 							 \\ 
      + \# of DB at 1000  &   \textbf{44.612} &66.798 & \textbf{49.652}&29.310&\textbf{ 52.751} & \textbf{57.443 }					 		 \\
      + NMS thresh at 0.6 &44.503 & 66.539 & 49.581& 29.264 & 52.635 & 57.228 							\\
      \hline
      
    \end{tabular}\vspace{-0.2cm}
    \captionof{table}{The table shows the final blending of all the modifications on top of the baseline. Here, the second and third rows of all table sections mean that the model mentioned on the first row was supplemented with setting the detection boxes at 1000 and then setting the NMS threshold at 0.6.}
  \label{tab:beast}
\end{minipage}
\end{table*}
}

\begin{figure}[htp!]
    \centering
    \includegraphics[width=0.35\textwidth,height=0.3\textheight,keepaspectratio]{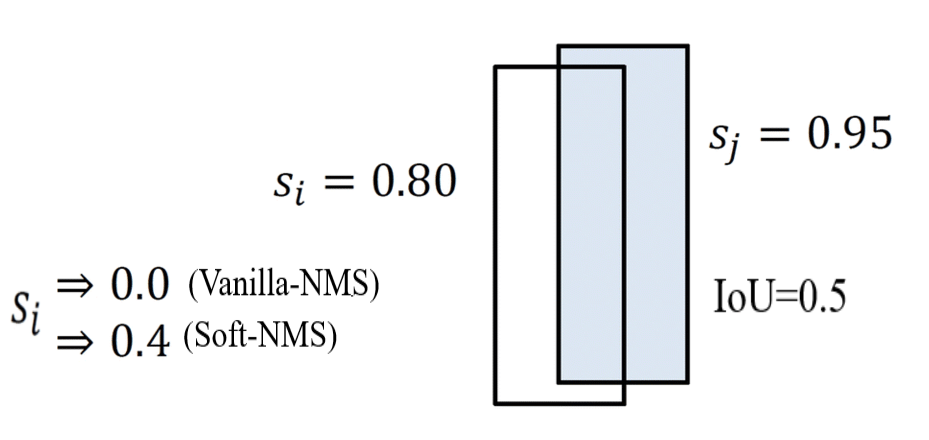}
    \caption{The figure shows the difference between soft-NMS and vanilla NMS with $N_t=0.5$. In case of vanilla-NMS the second bounding box($s_j$) is neglected, while in soft-NMS the bounding box is kept with lower score.}
    \label{soft}
\end{figure}

To address this issue, we opted to use Soft-NMS \cite{soft} rather than vanilla NMS in our project. As shown in Figure \ref{soft}, unlike vanilla NMS, Soft-NMS does not fully suppress other bounding boxes; instead, it decays the score as a continuous function in the IoU with the object as follows,
\begin{equation}
  s_j=\begin{cases}
    s_j, & \text{ $IoU > N_t$}.\\
    s_j(1-IoU), & \text{$IoU\leq N_t$}.
  \end{cases}
\end{equation}
In doing so, we avoid neglecting bounding boxes that actually contains an object but in close proximity to another object. This case occurs for multiple samples in our dataset.
\subsubsection{Changing Anchor sizes}
The default anchor sizes of $[[32], [64], [128], [256], [512]]$ chosen at each scale in the FPN based Faster R-CNN work well for relatively larger objects as compared to the objects in our dataset. In order for the model to work well on our dataset, we customize the size of the anchors based on the size distribution of the ground truth bounding boxes in our dataset.
From Figure \ref{data_analysis}, we can observe that most of the objects occupy around $15-20\%$ of the image area. Taking into account that most of the bounding boxes are small, we decrease the size of the anchors to cover the entire range of small objects in our dataset. 

\section{Experiments \& Results}
\paragraph{Evaluation Metrics.}
Standard COCO metrics \cite{mscoco}: AP (averaged over IoU threshold), $AP_{50}$, $AP_{75}$, $AP_S$, $AP_M$ and $AP_L$, where S,M and L represent small (area: 10-144 pixels), medium (area:144 to 1024 pixels) and large objects(area: 1024 and above), are used for evaluation.

\paragraph{Experimental Setup}
We carried out all the experiments with a batch size of $2$, and reduced the number of ROI proposals to be passed to ROI Head by a factor of $2$ (default is 512) due to the computational constraints of our setup. All of our experiments were trained on a 24GB NVIDIA Quadro RTX 6000 GPU. For finding the best data augmentation, the relevant experiments were trained for $60,000$ and a base learning rate of $0.00025$. Later on, all the models were trained for $100,000$ iteration with a base learning rate of $0.0025$, which is decreased by a factor of 10 at $50,000$ and $85,000$ iterations. In the experiments, Soft-NMS is utilised, a linearly decaying scoring function is used for reducing the score of detection boxes.

\paragraph{Results of Soft-NMS and Anchor Sizes.}
Since the most objects in our dataset are small in size, using small anchor sizes was hypothesized to increase the performance of model on small objects. This proved to be true, as shown in Table \ref{tab:nms}. Here, $A1$ represents [16, 64, 128, 256, 512] and $A2$ represents [8, 64, 128, 256, 512] anchor sizes. From the Table \ref{tab:nms}, we can observe that the model with smaller anchor sizes, $A1$ or $A2$, lead to a higher $AP_S$ score. However, we can also see that utilizing smaller anchor causes $AP_L$ score of larger objects to fall. To deal with this issue, Soft-NMS comes into play. Table \ref{tab:nms} also compares the performance of the baseline model with NMS and Soft-NMS under different anchor sizes. As can be seen, the mean average precision increased from 36.04 to 36.397 when we use Soft-NMS with default anchor sizes. Using the above anchor sizes ($A1$ and $A2$) improved the mAP even more, with $A2$ and Soft-NMS powered baseline yielding the highest mAP of 36.644. This combination provided the best $AP$ scores across almost all the categories. The experiments reported in Table \ref{tab:nms} were trained for $60,000$ iterations with a base learning rate of $0.00025$.

\paragraph{Results of Backbone Change.}
Changing solely the backbone from ResNet-101 to ResNetV2-101, which is supported with a smoother flow of gradients in the skip connections, group normalization, weight standardization and pretrained quality with ImageNet21k, showed a promising boost. Table \ref{tab:backbone} reveals that the mAP rose to 41.977 from the baseline, with all the other metrics tailing the same trend. The experiments reported in Table \ref{tab:backbone} were trained for $60,000$ iterations with a base learning rate of $0.0025$.

\paragraph{Results of Deep-RPN.}
Integrating the model RPN with spatial and channel squeeze and excitation blocks, as well as deepening it aided the model to learn more beneficial features, resulting in higher mAP. As can be seen in Table \ref{tab:deeprpn}, our baseline, empowered with ResNetv2, deep-RPN module and the smaller anchor sizes reached the mAP of 43.962 when the number of detection boxes was set at 1000 and NMS threshold was set to 0.6. A similar increasing pattern was observed with other metrics of mAP. The experiments reported in Table \ref{tab:deeprpn} were trained for $100,000$ iterations with a base learning rate of $0.0025$.

\paragraph{Results of Blends.}
Finally, putting all the tweaks onto one place, two versions of what we call blend model are proposed. In Table \ref{tab:beast}, BlendV1 is our baseline Faster R-CNN with FPN, strengthened with ResNetv2-101 backbone, deeper RPN, smaller anchor sizes ($A2$), and a set of data augmentations. Here, data augmentation set comprises resize shortest edge, horizontal flip, vertical flip and saturation, which provided the highest mAP values as reported in Supplementary Table \ref{tab:data_augs}. BlendV2 is the same model with an addition of Soft-NMS. BlendV1 computed at 1000 detection boxes with NMS threshold at 0.6 reached 44.457 mAP, and BlendV2 threholded at 1000 detection boxes landed at the highest of 44.612 mAP. Note that the small objects precision was high with BlendV1, however, medium and large object suffered a small decrease. This issue was resolved with BlendV2 with the help of Soft-NMS.

\begin{figure}[h!]
    \centering
    \includegraphics[width=0.45\textwidth]{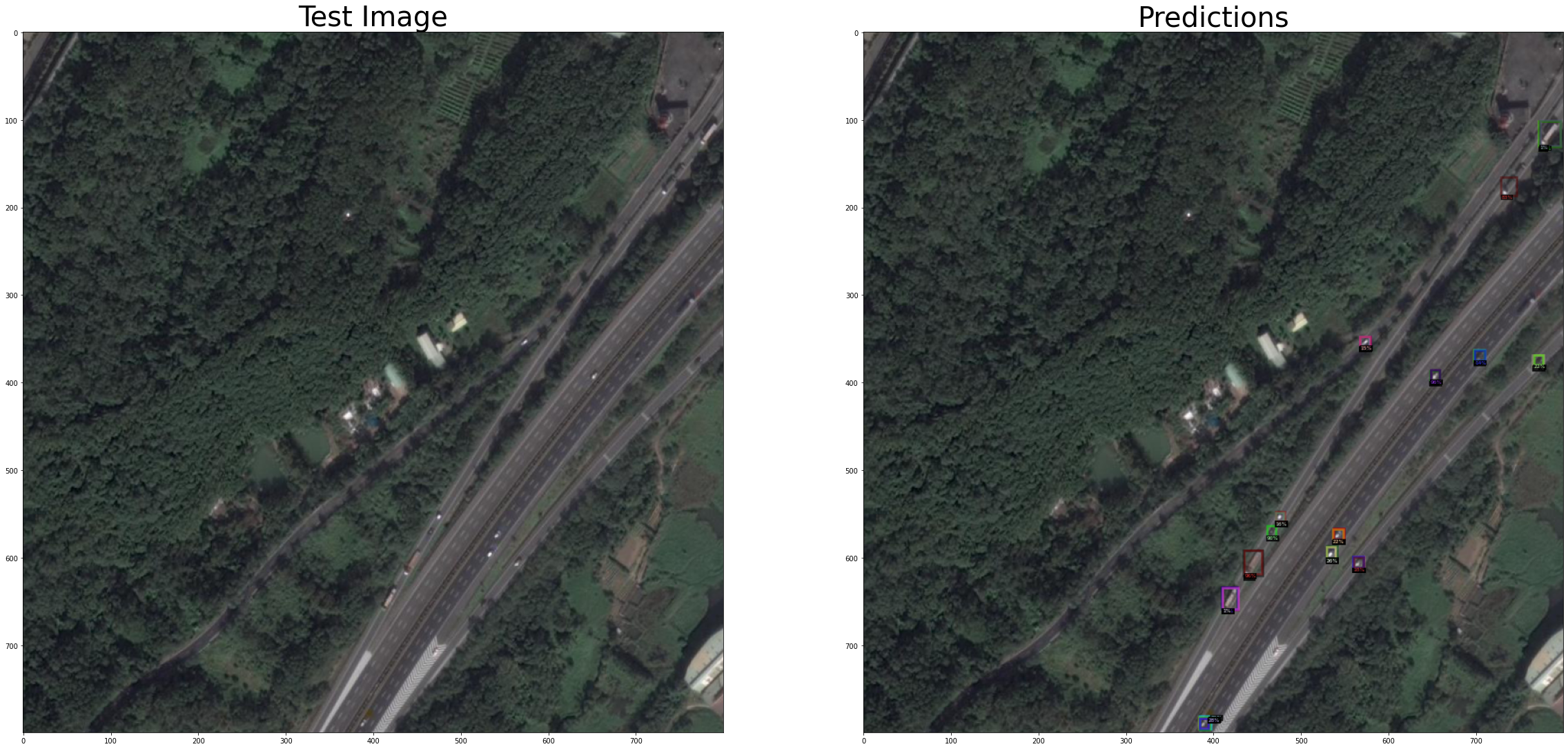}
    \caption{The figure shows output from BlendV2.}
    \label{pred1}
\end{figure}

\begin{figure}[h!]
    \centering
    \includegraphics[width=0.45\textwidth]{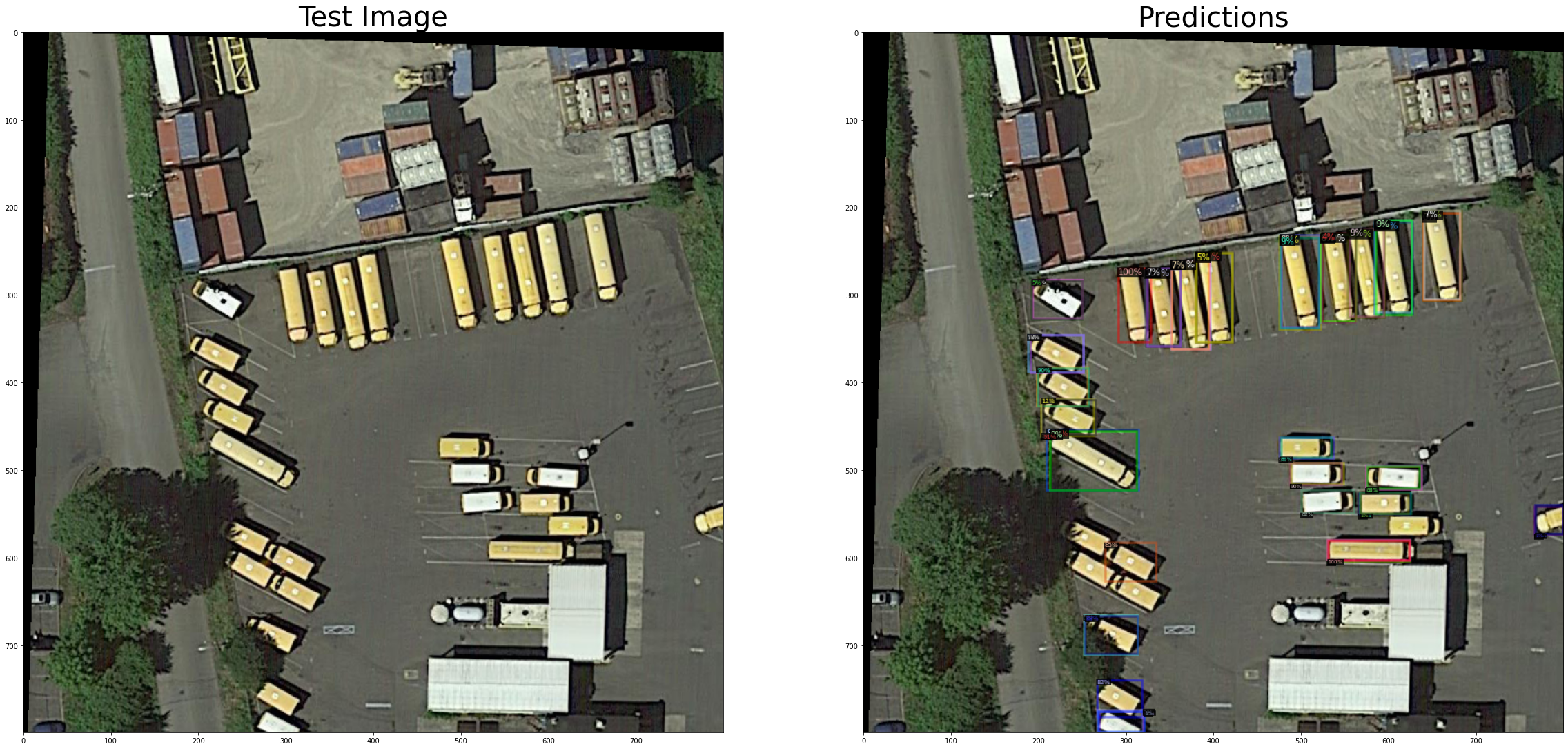}
    \caption{The figure shows output from BlendV2.}
    \label{pred2}
\end{figure}

\begin{figure}[t!]
    \centering
    \includegraphics[width=0.45\textwidth]{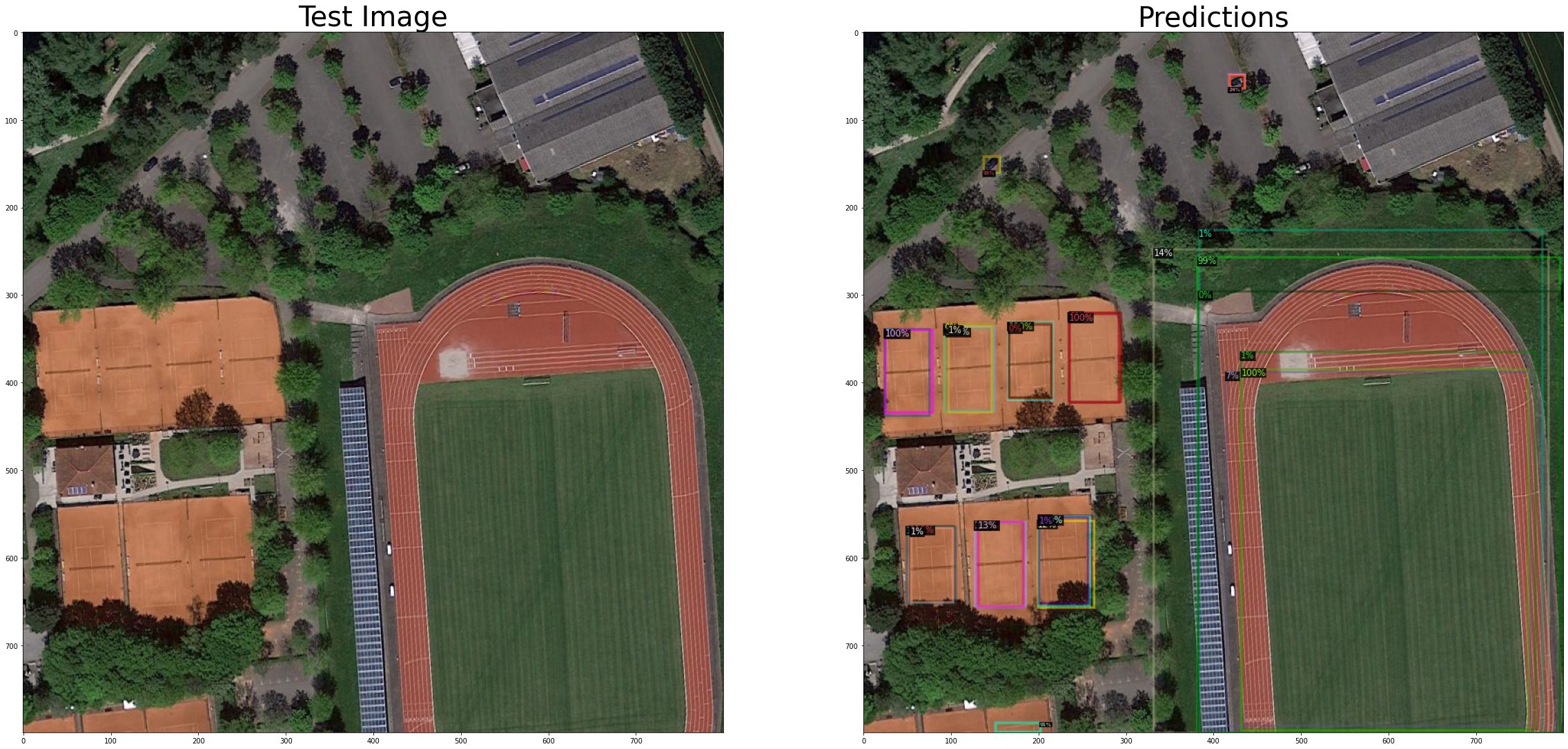}
    \caption{The figure shows output from BlendV2.}
    \label{pred3}
\end{figure}

\paragraph{Qualitative Results.}
The Figures \ref{pred1}, \ref{pred2}, and \ref{pred3} show the original image on the left and predicted detection boxes drawn on them on the right. As can be seen, the model outputs detection boxes even when the objects are small very accurately.

\section{Discussion and Conclusion}
This paper presents a set of techniques that can be applied to boost the network, Faster R-CNN in particular, when trained for aerial images. With an extensive analysis, we show that the baseline model can dress up with certain additional features such as the backbone change, data augmentations, SE-based deeper RPN and soft NMS to reach higher results. Such a huge jump in the mAP proves that the integration of such components to the network can benefit it to a great extent. Utilizing spatial and channel squeeze and excitation block in the RPN as well as increasing the layers of the RPN are considered as our contributions of the work, as a cherry on top of all the other feature combinations. The loss function, an integral part of the neural networks, can be of tremendous help too, if properly chosen and tuned. In our case, the loss functions we experimented with (as described in details in Supplementary section) did not contribute to the success of the network. We believe it is due to the fact that the loss functions were not fine-tuned for the current task, and with more experiments, a reasonable improvement may be achieved. We hope to further investigate a better classification and regression loss suited for countering the challenges of significant scale variation and the class imbalance problem prevalent in the task.

\newpage
\bibliography{egbib}

\begin{table*}\centering
  \small
\setlength{\tabcolsep}{4.5pt}
    \begin{tabular}{l|ccc|ccc}
      Augmentations & AP & $AP_{50}$ & $AP_{75}$ & $AP_{S}$ & $AP_{M}$ & $AP_{L}$  \\ %
      \hline
      Resize SE, HFlip (baseline)  & 35.284 & 56.698 & 37.901 & 20.609 & 42.403 & 46.621  \\ 
      
      \hline
      Resize SE, HFlip, VFlip  & 35.457 & 56.761 & 38.488 & 20.481 & 42.433 & 47.864  \\
      \hline
      Resize SE, HFlip, Rotation(90)  & 33.679 & 53.569 & 36.644 & 19.733 & 40.206 & 45.522  \\
      \hline
      Resize SE, HFlip, VFlip, Brightness & 35.356 & 56.596 & 38.641 & 20.797 & 42.146 & 47.303  \\
      \hline
      Resize SE, HFlip, VFlip, Saturation  & \textbf{36.181} & \textbf{56.967} & \textbf{38.736} & \textbf{21.420} & \textbf{42.979} & \textbf{48.428}   \\
      \hline
    \end{tabular}\vspace{-0.2cm}
    \captionof{table}{The table shows the results obtained from using various data augmentations.}
  \label{tab:data_augs}\vspace{0.1cm}
\end{table*}

\begin{table*}\centering
  \small
\begin{minipage}{0.65\textwidth}
    \begin{tabular}{l|ccc|ccc}
      Method & AP & $AP_{50}$ & $AP_{75}$ & $AP_{S}$ & $AP_{M}$ & $AP_{L}$  \\ %
      \hline
      Baseline-ResNetv2 & 43.06	&65.715	&47.52&	28.3&	51.143&	56.81 \\ 
      + \# of DB at 1000  &  43.414&	66.409&	47.885&	28.512&	51.549&	\textbf{57.275 }\\
      + NMS thresh at 0.6 & 43.466&	66.282&	48.021&	28.561&	\textbf{51.618}&	57.22\\
      \hline
      Baseline-ResNetv2+A2 & 43.085	&66.34&	47.585&	28.787&	49.962&	55.218 \\ 
      + \# of DB at 1000  &  43.715	&\textbf{67.675}&	48.118&	29.33&	50.572&	55.853 \\
      + NMS thresh at 0.6 &\textbf{43.717}&	67.432&	\textbf{48.261}&	\textbf{29.346}&	50.558&	55.657\\
      \hline

    \end{tabular}\vspace{-0.2cm}
    \caption{The table shows the results of thresholding detection boxes at 1000 and NMS at 0.6.}
  \label{tab:anchoring}\vspace{-0.1cm}
\end{minipage}
\end{table*}

\newpage
\section{Supplementary Materials}
\subsection{Further Analysis on Dataset}
\begin{figure}[h!]
    \centering
    \includegraphics[width=0.45\textwidth]{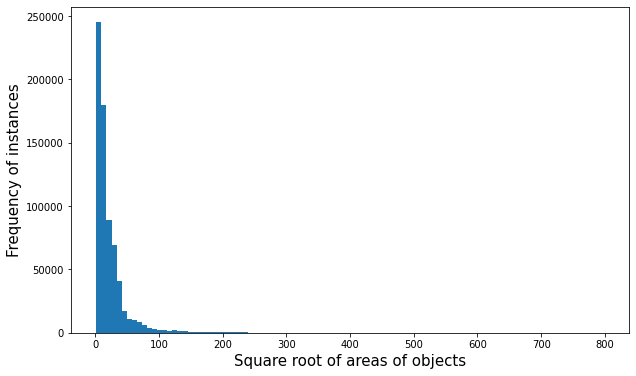}
    \caption{The figure shows frequencies of instances (y-axis) having the square root of areas of objects (x-axis).}
    \label{sqrt}
\end{figure}

\begin{figure}[h!]
    \centering
    \includegraphics[width=0.09\textwidth]{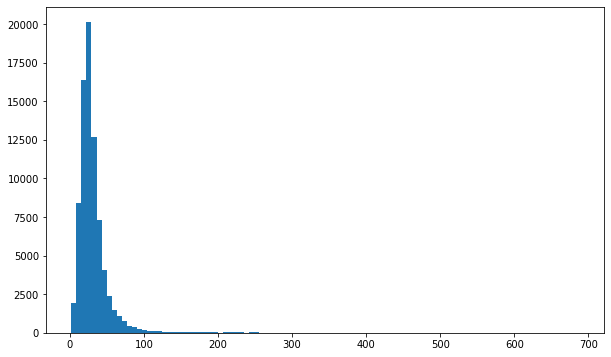}
    \includegraphics[width=0.09\textwidth]{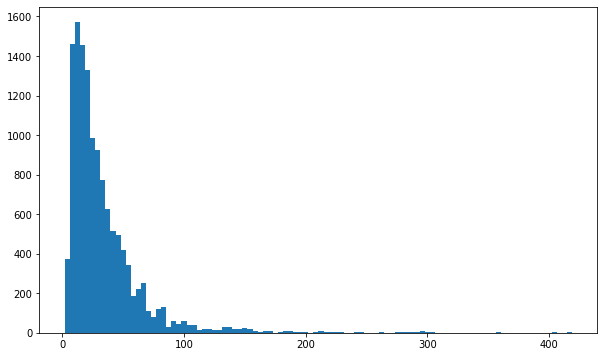}
    \includegraphics[width=0.09\textwidth]{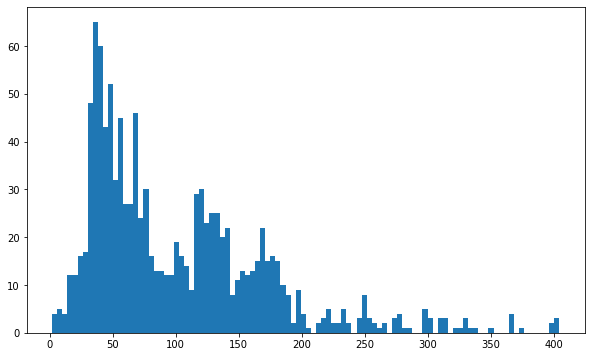}
    \includegraphics[width=0.09\textwidth]{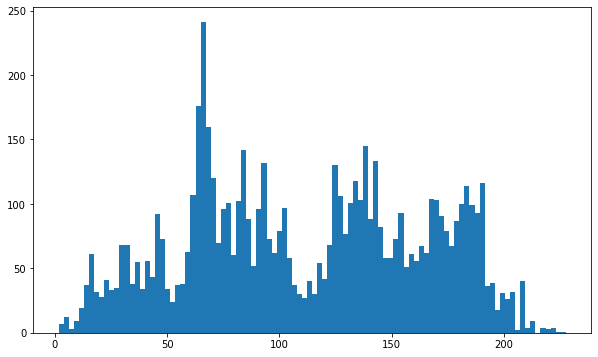}
    \includegraphics[width=0.09\textwidth]{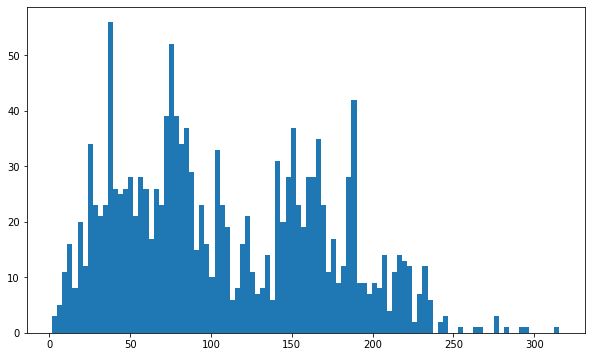}
    \includegraphics[width=0.09\textwidth]{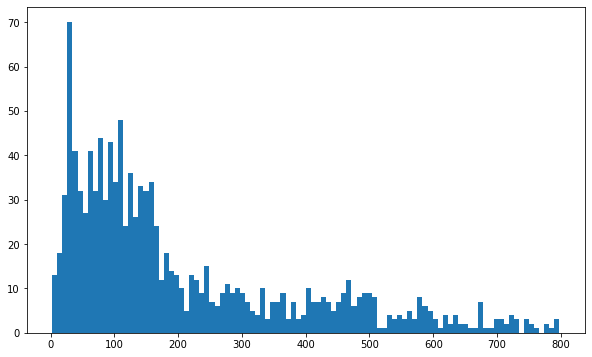}
    \includegraphics[width=0.09\textwidth]{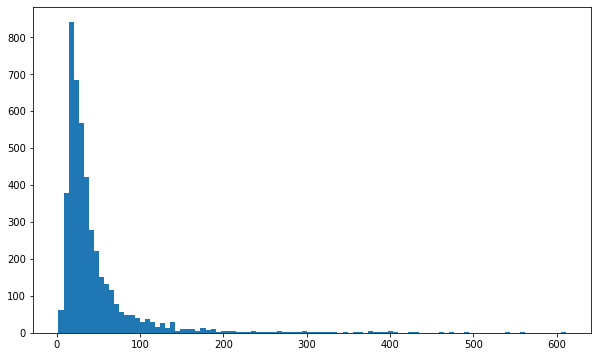}
    \includegraphics[width=0.09\textwidth]{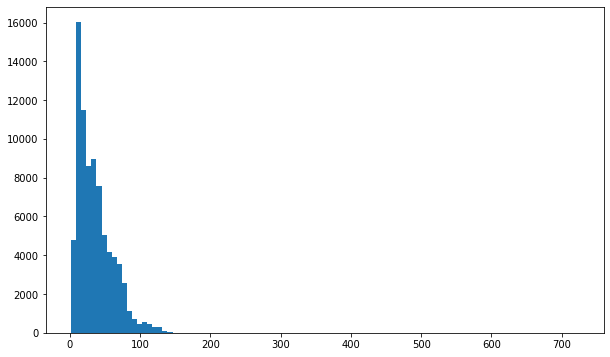}
    \includegraphics[width=0.09\textwidth]{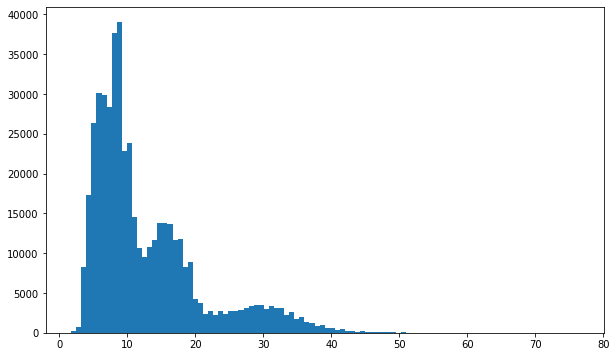}
    \includegraphics[width=0.09\textwidth]{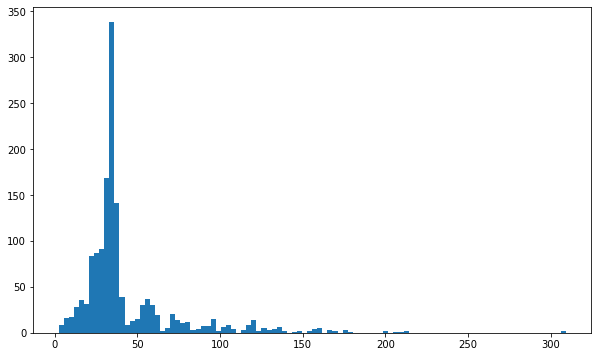}
    \includegraphics[width=0.09\textwidth]{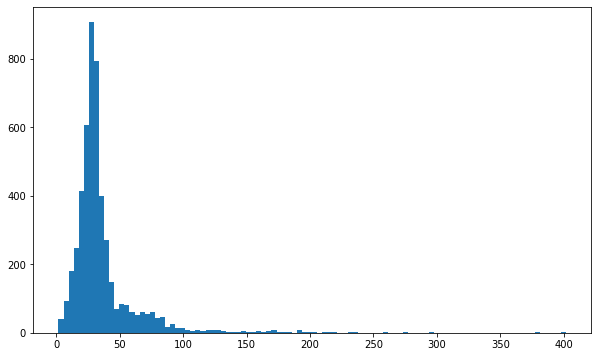}
    \includegraphics[width=0.09\textwidth]{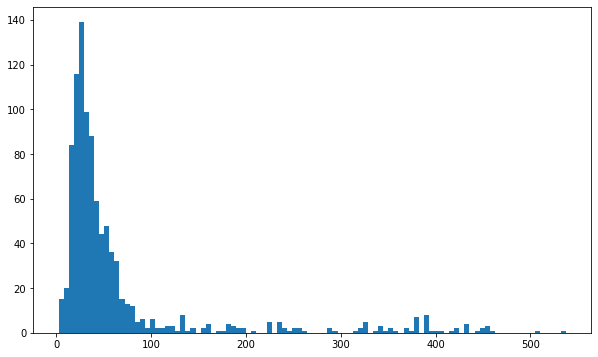}
    \includegraphics[width=0.09\textwidth]{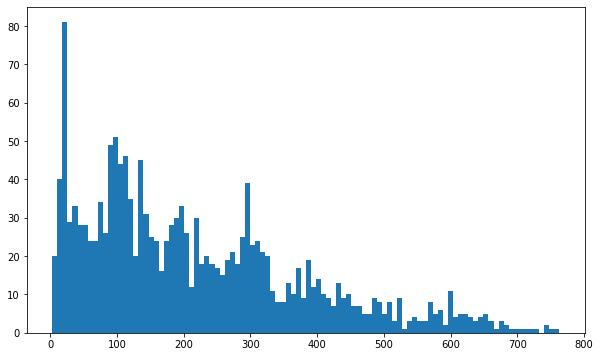}
    \includegraphics[width=0.09\textwidth]{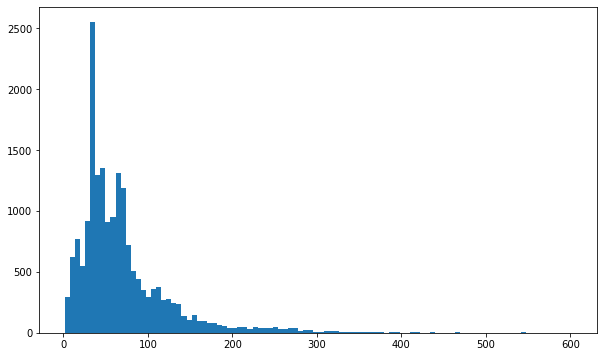}
    \includegraphics[width=0.09\textwidth]{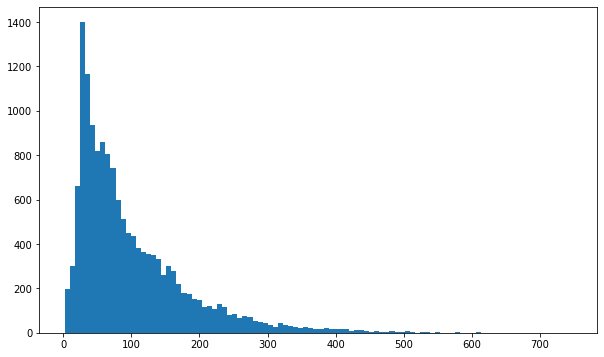}

    \caption{The figure shows frequencies of instances (y-axis) having the square root of areas of objects (x-axis) for each category.}
    \label{sqrt_cat}
\end{figure}

To further investigate the dataset, the areas of object instances are calculated and the square root of them are derived to understand the average frequency of such areas in different categories. Figure \ref{sqrt} shows the frequency of instances having similar areas of objects, indicating that the majority of objects have a square root of an area under 100. This is used to give a sense of measure to the dimensions of objects and it is useful to provide information on what sizes of proposals can be utilized when training. Similarly, the calculations are done for each category to see the range of different size variations, as depicted in Figure \ref{sqrt_cat}. Note that the axes have different ranges here as this was the only noticeable way to plot them in a comparative manner. The same pattern is shown here that each category holds the highest count of square root of areas that are under around 100. 

\subsection{Detailed Experiments \& Results}

\subsubsection{Loss Functions}
\begin{table*}\centering
  \small
\begin{minipage}{0.65\textwidth}
    \begin{tabular}{l|ccc|ccc}
      Method & AP & $AP_{50}$ & $AP_{75}$ & $AP_{S}$ & $AP_{M}$ & $AP_{L}$  \\ %
      \hline
      Baseline-ResNetv2-50 & 42.617 & 64.627 & 47.255 & 28.033 & 50.448 & 56.311 \\ 
      + \# of DB at 1000  &  43.063 & 65.294 & 47.748 & 28.356 & 51.002 & 56.954 \\
      + NMS thresh at 0.6 & 42.970 & 65.023 & 47.708 & 28.306 & 50.897 & 56.724 \\
      \hline
 
    \end{tabular}\vspace{-0.2cm}
    \caption{The table shows the results obtained by changing the backbone from ResNetv2-101 to ResNetv2-50.}
  \label{tab:resnet50}\vspace{0.1cm}
\end{minipage}
\end{table*}

\begin{table*}\centering
  \small
\begin{minipage}{0.65\textwidth}
    \begin{tabular}{l|ccc|ccc}
      Method & AP & $AP_{50}$ & $AP_{75}$ & $AP_{S}$ & $AP_{M}$ & $AP_{L}$  \\ %
      \hline
      Baseline-ResNetv2 & 43.06	&65.715	&47.52&	28.3&	51.143&	56.81 \\ 
      + \# of DB at 1000  &  43.414&	66.409&	47.885&	28.512&	51.549&	57.275 \\
      + NMS thresh at 0.6 & \textbf{43.466}&	\textbf{66.282}&	\textbf{48.021}&	\textbf{28.561}&	\textbf{51.618}&\textbf{	57.22}\\
      \hline
      Baseline-ResNetv2+OPL & 43.038&	65.521&	47.308&	27.954&	50.861&	56 \\ 
      + \# of DB at 1000  &  43.371	&66.177&	47.635&	28.162&	51.284&	56.464 \\
      + NMS thresh at 0.6 &43.389&	66.01&	47.702&	28.184&	51.363&	56.328\\
      \hline
      \hline
      Baseline-ResNet  & 36.099 & 57.259 & 38.934 & 21.256 & 43.518 & 49.393  \\ 
      \hline
      ACLS (0.7 threshold) & 29.162 & 46.907 & 31.041 & 17.645 & 35.193 & 39.933    \\
      \hline
      ACLS (0.5 threshold)  & 30.588 & 49.475 & 32.689 & 18.105 & 37.622 & 41.683   \\
      \hline
      ACLS (0.4 threshold) & 31.946 & 51.217 & 34.095 & 18.725 & 38.929 & 43.583   \\
      \hline
      ACLS (0.3 threshold)  & 32.123 & 51.628 & 34.273 & 18.508 & 39.081 & 44.197    \\
      \hline
    \end{tabular}\vspace{-0.2cm}
    \captionof{table}{The table shows the results of using Orthogonal Projection Loss (OPL) and Adaptive Class Suppression Loss (ACSL). OPL results were evaluated on ResNetv2 backbone, while ResNet backbone was used in ACLS experiments.}
  \label{tab:opl}\vspace{-0.1cm}
\end{minipage}
\end{table*}

\paragraph{Orthogonal Projection Loss.} Recently, a new loss function was proposed for the task of classification, known as Orthogonal Projection Loss (OPL)\cite{OLS}. This loss function complements the cross entropy loss in having well separated features for different classes. OPL imposes orthogonality constraints on the mini-batch level, forcing inter-class separation as well as intra-class clustering. We use this objective function for the purposing of aiming a better discrimination between the background and foreground classes which can further enhance the prediction of our model.

Consider $F$ be a deep neural network, which can be separated into $F_{\psi}$ and $F_{\phi}$ which are the feature extraction block and the classification head respectively. Given an input-output pair $\{x_i, y_i\}$, $f_{i} =F_{\psi}(x_i) $ will be the intermediate features of the network. The OPL enforces the features belonging to different classes to be orthogonal to each other, while features from the same class should be similar. The resulting loss function on a batch of images $B$ is :

\begin{equation}
    s=\sum_{i, j \in B \atop y_{i}=y_{j}}< f_{i}, f_{j}>
\end{equation}

\begin{equation}
    d=\sum_{i, k \in B \atop y_i \neq y_{k}} <f_{i}, f_{k}>
\end{equation}

\begin{align}
    L_{OPL} = (1-s) + |d|
\end{align}

where $<.,.>$ is the cosine similarity operator and $|.|$ is the absolute value operator. So, our overall loss classification loss is the combination of CE and OPL loss:
\begin{align}
    L = L_{CE} + \lambda*L_{OPL}
\end{align}
where $\lambda$ is a hyper-parameter to control the effect of the OPL loss.
    
\paragraph{OPL Results.}
Table \ref{tab:opl} shows the results of baseline with ResNetv2 with and without OPL loss. As is evident, the introduced loss function did not perform well in any part of the metric. This can be on account of the small object sizes and overlapping feature our dataset holds. Although the loss function is capable of discriminating between fore- and background objects from each other, small area coverage by objects and them overlapping each other in many cases may be causing the objective function confuse, therefore, yield unsatisfactory results. 

\paragraph{Adaptive Class Suppression Loss.}
In order to deal with the class imbalance in our dataset, which is known as long tail problem, we replaced the vanilla cross-entropy loss with Adaptive Class Suppression Loss (ACSL)\cite{acsl}. Cross-entropy goal is to output a hot one confidence vector with one at the correct class prediction and zero otherwise. To accomplish this, it trains the classifier to output negative suppression gradients for negative class to output low confidence for this classes, this would be useful in case of a balanced dataset, but in case of long tail datase, some classes rarely appear so the model will learn to always suppress them. ACSL addresses this problem , as shown in Eq.\ref{loss} by adding a parameter $w$ to only output negative suppressing gradients for the negative classes that causes confusion, i.e. have high confidence, and does not suppress other negative classes. 
\begin{equation}
   L_{ACSL}(x_s)=-\sum_{i=1}^C w_i log(p_i)
   \label{loss}
\end{equation}
In case of the positive class, $w_i$ is 1 so the positive gradient flows normally. In case of the negative class $w_i$ is set to zero if the confidence is lower than a threshold $\eta$ to negate the suppression gradient, otherwise it is set to 1. This is expressed in the following equation,
\begin{equation}
  w_i=\begin{cases}
   1, & \text{if $i=k$ }\\
   1, & \text{if $i \neq k$ and $p_i \geq \eta$ }\\
   0, & \text{if $i \neq k$ and $p_i < \eta$ } .
  \end{cases}
\end{equation}

Also classes are split into three groups, according to the number of samples, which are:rare,common and frequent, and percentage of background samples are suppressed in case of rare and common classes. Only 1\% of background samples are taken into account in ca  se of rare class and 10\% in case of common class.
\paragraph{ACSL Results.}
Table \ref{tab:opl} shows the results of baseline with ResNet101 with and without ACSL loss. As shown, the introduced loss function did not perform well in any part of the metric. This can be on account for inaccurate hyper-parameters or split of classes which required further analysis that the time did not allow.
\subsubsection{New Backbone}
To make further comparisons on the backbone of the Faster R-CNN network, instead of using ResNetv2-101, we experimented with ResNetv2-50 as a backbone to see how much it would affect in terms of accuracy. To perform this experiment, we are using our BlendV2, the best performing model, only changing the backbone to ResNetv2-50. Table \ref{tab:resnet50} shows that the results suffered a small decrease with the introduction of the ResNetv2-50, a smaller network. It proves that the model backbone we chose for BlendV1 and V2 are crucial for the whole network to reach such a score. 

\end{document}